\documentclass[Afour,sageh,times]{sagej}

\usepackage{moreverb,url}

\usepackage[colorlinks,bookmarksopen,bookmarksnumbered,citecolor=red,urlcolor=red]{hyperref}

\newcommand\BibTeX{{\rmfamily B\kern-.05em \textsc{i\kern-.025em b}\kern-.08em
T\kern-.1667em\lower.7ex\hbox{E}\kern-.125emX}}

\def\volumeyear{2016}

\usepackage{hyperref}
\hypersetup{
    colorlinks=true,
    linkcolor=red,
    filecolor=magenta,
    urlcolor=red
}

\usepackage{esvect}
\usepackage[percent]{overpic}
\usepackage{tabu}
\usepackage{multirow}
\def\mr{\multirow}
\def\mc{\mc}

\newcommand{\ul}[1]{\underline{#1}}
\newcommand{\tb}[1]{\textbf{#1}}

\usepackage{balance}

\usepackage{subcaption}
\usepackage{caption}

\usepackage[overload]{empheq}

\DeclareMathOperator*{\argmin}{arg\,min}

\newcommand{\norm}[1]{\left\lVert#1\right\rVert}
\def\wrt{\text{w.r.t. }}

\newcommand{\fr}[1]{\texttt{#1}}
\def\fW{\fr{W}}
\def\fA{\fr{A}}
\def\fB{\fr{B}}

\def\fP{\fr{P}}
\def\fL{\fr{L}}
\def\fC{\fr{C}}
\def\pos{\mathbf{p}}
\def\rot{\mathbf{R}}
\def\tf{\mathbf{T}}
\def\trans{\mathbf{t}}

\newcommand{\highlight}[1]{#1}

\begin{document}

\runninghead{NTU VIRAL DATASET}

\title{A demonstration of the \LaTeXe\ class file for
\itshape{SAGE Publications}}
\title{NTU VIRAL: A visual-inertial-ranging-lidar dataset, from an aerial vehicle viewpoint}

\author{Thien-Minh Nguyen\affilnum{1},
Shenghai Yuan\affilnum{1},
Muqing Cao\affilnum{1},
Yang Lyu\affilnum{1},
Thien Hoang Nguyen\affilnum{1},
Lihua Xie\affilnum{1}}

\affiliation{\affilnum{1}Nanyang Technological University, Singapore}

\corrauth{Lihua Xie, School of Electrical and Electronic Engineering, Nanyang Technological University, 50 Nanyang Ave, Singapore, 639798}

\email{lhxie@ntu.edu.sg}

\begin{abstract}
In recent years, autonomous robots have become ubiquitous in research and daily life. Among many factors, public datasets play an important role in the progress of this field, as they waive the tall order of initial investment in hardware and manpower. However, for research on autonomous aerial systems, there appears to be a relative lack of public datasets on par with those used for autonomous driving and ground robots. Thus, to fill in this gap, we conduct a data collection exercise on an aerial platform equipped with an extensive and unique set of sensors:
two 3D lidars, two hardware-synchronized global-shutter cameras, multiple Inertial Measurement Units (IMUs), and especially, multiple Ultra-wideband (UWB) ranging units. The comprehensive sensor suite resembles that of an autonomous driving car, but features distinct and challenging characteristics of aerial operations. We record multiple datasets in several challenging indoor and outdoor conditions. Calibration results and ground truth from a high-accuracy laser tracker are also included in each package. All resources can be accessed via our webpage \url{https://ntu-aris.github.io/ntu_viral_dataset/}.
\end{abstract}

\keywords{Dataset, Aerial Robot, Autonomous System, Simultaneous Localization and Mapping}

\maketitle
%

\renewcommand{\thefootnote}{\arabic{footnote}}

\section{Introduction}

Over the years, autonomous systems have made significant progress.
One of the most crucial factors for these advancements can be attributed to the public data suites.
On one hand, these datasets can waive the inhibitive requirements on budget and manpower, e.g. hardware development, calibration, field operations, etc. for individual researchers. Hence, researchers can easily investigate new navigation schemes and put them to test across a variety of scenarios and environments. On the other hand, the benchmark tools also help streamline the verification, evaluation and comparison between proposed methods, thus allowing navigation methods to be fairly rated and ranked based on a common basis.

\begin{figure}[t]
    \centering
    \includegraphics[width=\linewidth]{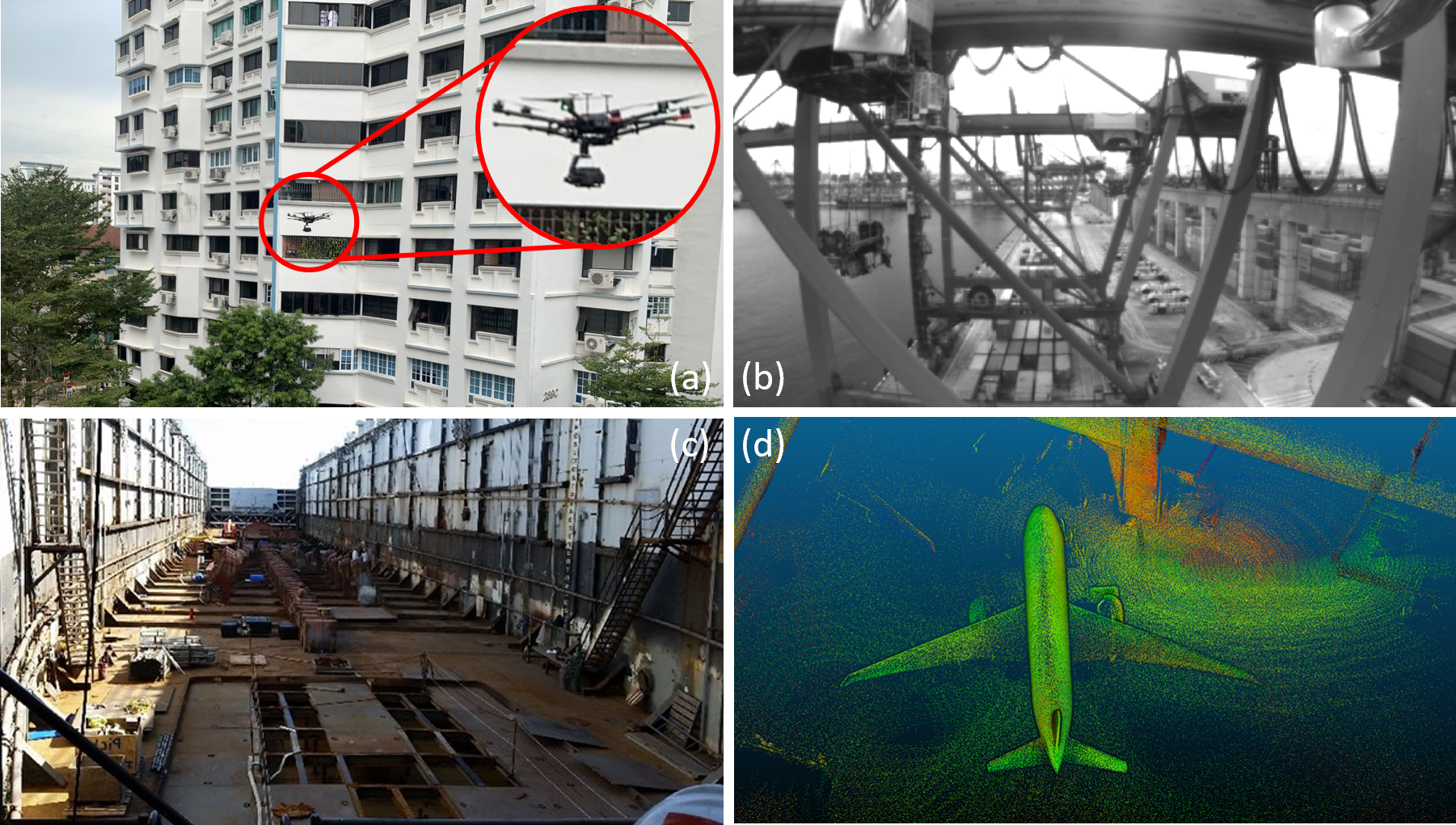}
	\caption{Typical inspection applications for autonomous AVs in GPS-denied environments: (a) building complex, (b) cranes, (c) cargo ships (d) hangar}
	\label{fig: environments}
\end{figure}

Indeed, there are many datasets for a variety of tasks on autonomous vehicles, e.g. object recognition, stereo depth perception, scene understanding, traffic analysis, etc. In this paper we only focus on the datasets that can be used to investigate navigation capability of aerial vehicles (AVs) in GPS-denied environments, 
Fig. \ref{fig: environments} illustrates a range of such environments where the application of autonomous AV system is being investigated.
We review some of the datasets that are deemed most relevant to our targeted applications above. Table \ref{tab: datasets and sensors} gives a summary of these datasets and their features, to the best of our knowledge.

\subsection{Related works: the ground-air dichotomy}

\begin{table*}[t]
\renewcommand{\arraystretch}{1.25}

\caption{Notable public datasets divided by their ground and air focuses.}
\label{tab: datasets and sensors}
\resizebox{\linewidth}{!}
{%
\centering
\begin{tabular}{lllllll} 
\hline\hline
\multicolumn{1}{c}{\multirow{2}{*}{ \textbf{Dataset} }}
    & \multicolumn{4}{c}{\textbf{Sensor used} }
    & \multicolumn{1}{c}{\multirow{2}{*}{\textbf{Ground truth} }}
    & \multicolumn{1}{c}{\multirow{2}{*}{\begin{tabular}[c]{@{}c@{}}\textbf{Mobile}\\\textbf{Platform} \end{tabular}}}
\\\cline{2-5}

\multicolumn{1}{c}{}
    & \multicolumn{1}{c}{\textbf{IMU} }
    & \multicolumn{1}{c}{\textbf{Camera} }
    & \multicolumn{1}{c}{\textbf{Lidar} }
    & \multicolumn{1}{c}{\textbf{UWB} }
    & \multicolumn{1}{c}{}
    & \multicolumn{1}{c}{}
\\\hline

\begin{tabular}[c]{@{}l@{}}MIT DARPA,\\\cite{huang2010high}\end{tabular}
    & N/A
    & \begin{tabular}[c]{@{}l@{}}4 Point Grey: 4×376×240 @10Hz\\ 1 Point Grey: 752×480 @22.8Hz \end{tabular}
    & \begin{tabular}[c]{@{}l@{}}3D Velodyne HDL-64E @15Hz\\ 12 2D-SICK @75Hz \end{tabular}
    & N/A
    & GPS/INS
    & Car
\\\hline

\begin{tabular}[c]{@{}l@{}}Ford Campus,\\\cite{pandey2011ford}\end{tabular}
    & acc/gyr @100Hz
    & LadyBug 3: 6×1600×600 @8Hz
    & \begin{tabular}[c]{@{}l@{}}3D-Velodyne HDL-64E @10Hz\\ 2 2D-Riegl LMS \end{tabular}
    & N/A
    & GPS/INS
    & Car
\\\hline

\begin{tabular}[c]{@{}l@{}}KITTI,\\\cite{Geiger2013IJRR}\end{tabular}
    & acc/gyr @10Hz
    & \begin{tabular}[c]{@{}l@{}}2 Point Grey (gray): 2×1392×512 @10Hz\\ 2 Point Grey (color): 2×1392×512 @10Hz \end{tabular}
    & 3D-Velodyne HDL-64E @10Hz
    & N/A
    & RTK GPS/INS
    & Car
\\\hline

\begin{tabular}[c]{@{}l@{}}NCLT,\\\cite{carlevaris2016university}\end{tabular}
    & acc/gyr @100Hz
    & LadyBug 3: 6×1600×1200 @5Hz
    & \begin{tabular}[c]{@{}l@{}}3D-Velodyne HDL-32E @10Hz\\ 2 2D-Hokuyo @10-40Hz \end{tabular}
    & N/A
    & \begin{tabular}[c]{@{}l@{}}RTK-GPS\\ LIDAR SLAM \end{tabular}
    & \begin{tabular}[c]{@{}l@{}}Wheeled\\Robot \end{tabular}
\\\hline

\begin{tabular}[c]{@{}l@{}}Oxford RobotCar,\\\cite{maddern20171}\end{tabular}
    & acc/gyr @50Hz
    & \begin{tabular}[c]{@{}l@{}}BumbleBee: 2×1280×960 @16Hz\\ 3 Grasshoper2: 3×1024×1024 @11.1Hz \end{tabular}
    & \begin{tabular}[c]{@{}l@{}}2 2D-SICK @50Hz\\ 3D-SICK @12.5Hz \end{tabular}
    & N/A
    & GPS/INS
    & Car
\\\hline

\begin{tabular}[c]{@{}l@{}}KAIST Urban,\\\cite{jeong2019complex}\end{tabular}
    & \begin{tabular}[c]{@{}l@{}}acc/gyr @200Hz\\ FOG @1000Hz \end{tabular} & FLIR (color): 2×1280×560 @10Hz
    & \begin{tabular}[c]{@{}l@{}}2 3D-Velodyne-16 @10Hz\\ 2 2D-SICK @100Hz \end{tabular}
    & N/A
    & SLAM
    & Car
\\\hline

\begin{tabular}[c]{@{}l@{}}Newer College,\\\cite{ramezani2020newer}\end{tabular}
    & acc/gyr @650Hz
    & D435i (Infrared): 2×848×480 @30Hz
    & 3D-Ouster-64 @10Hz
    & N/A
    & 6DOF ICP
    & Handheld
\\\hline\hline

\textbf{NTU VIRAL (Ours)}
    & \begin{tabular}[c]{@{}l@{}}\textbf{acc/gyr/mag}\\\textbf{@385Hz} \end{tabular}
    & \textbf{IDS (gray): 2×752×480@10Hz,}
    & \textbf{2 3D-Ouster-16 @10Hz}
    & \begin{tabular}[c]{@{}l@{}}\textbf{4 on UAV}\\\textbf{3 anchors} \end{tabular}
    & \begin{tabular}[c]{@{}l@{}}\textbf{3D Laser}\\\textbf{Tracker} \end{tabular}
    & \textbf{UAV}
\\\hline\hline

\begin{tabular}[c]{@{}l@{}}UMA-VI,\\\cite{zuniga2020vi}\end{tabular}
    & acc/gyr @250Hz
    & \begin{tabular}[c]{@{}l@{}}BumbleBee: 2×1024×768 @12.5Hz\\IDS (gray): 2×752×480 @25Hz, \end{tabular}
    & N/A
    & N/A
    & SLAM
    & Handheld
\\\hline

\begin{tabular}[c]{@{}l@{}}UZH FPV,\\\cite{delmerico2019we}\end{tabular}
    & \begin{tabular}[c]{@{}l@{}}acc/gyr\\ @500/1000Hz\end{tabular}
    & \begin{tabular}[c]{@{}l@{}}Fisheye stereo: 2×640×480 @30Hz\\mDAVIS: 346×260 + event @50Hz \end{tabular}
    & N/A
    & N/A
    & \begin{tabular}[c]{@{}l@{}}\textbf{3D Laser}\\\textbf{Tracker} \end{tabular}
    & UAV
\\\hline

\begin{tabular}[c]{@{}l@{}}TUM VI,\\\cite{schubert2018tum}\end{tabular}
    & acc/gyr @200Hz
    & IDS (gray): 2×1024×1024 @20Hz
    & N/A
    & N/A
    & 6DOF MoCap
    & Handheld
\\\hline

\begin{tabular}[c]{@{}l@{}}Upenn Fast Flight,\\\cite{sun2018robust}\end{tabular}
    & acc/gyr @200Hz
    & FLIR (gray): 2×960×800 @40Hz
    & N/A
    & N/A
    & GPS
    & UAV
\\\hline

\begin{tabular}[c]{@{}l@{}}Zurich Urban MAV,\\\cite{majdik2017zurich}\end{tabular}
    & acc/gyr @10Hz
    & GoPro (color): 1920×1080 @30Hz
    & N/A
    & N/A
    & \begin{tabular}[c]{@{}l@{}}Aerial\\Photogrammetry \end{tabular}
    & UAV
\\\hline

\begin{tabular}[c]{@{}l@{}}EuRoC,\\\cite{burri2016euroc}\end{tabular}
    & acc/gyr @200Hz
    & 2 MT9V034: 2×752×480 @20Hz
    & N/A
    & N/A
    & \begin{tabular}[c]{@{}l@{}}6DOF MoCap /\\ 3D Laser\\Tracker \end{tabular}
    & UAV
\\\hline\hline
\end{tabular}

}

\end{table*}

From Table \ref{tab: datasets and sensors}, it is clear that there exist two distinct groups of datasets. For those collected from cars or ground robots, a large number of sensors is often present, especially lidar. For example, the MIT DARPA dataset is comprised of 12 2D lidars and 1 64-channel lidar, along with a significant number of cameras. Over the years, there appears to be a decline of interest in 2D lidar (whose role has been gradually replaced by 3D lidar), along with a reduced number of cameras. This reflects the trajectory of autonomous driving research in the last decade, where in the beginning sensors had limited capability and variety, thus quality was compensated by quantity. As the field progresses, sensors have become much more compact and efficient, along with better software and algorithms.

For aerial platforms, also referred to as \textit{Unmanned Aerial Vehicles (UAVs)}, \textit{Micro Aerial Vehicles (MAVs)}, \textit{aerial robots} or \textit{drones} in the literature, we can see a clear contrast to the ground-based datasets. In this case, due to the payload constraint, 3D lidar is often left out of consideration. Instead, the focus is often on high frame-rate camera systems for self-localization capability at high speed and aggressive maneuvers, often under low lighting conditions, with the goal of pushing the capability of Visual-Inertial Odometry (VIO) and vision-based Simultaneous Localization and Mapping (SLAM) systems to cover ever more extreme conditions.

It can be seen that the dataset we are presenting in this paper falls in the middle of the two aforementioned classes.
For those datasets using cars and wheeled robots, the vehicle's trajectory is often close to the ground plane, without much abrupt translation or rotation motions. This is an advantage that has been well exploited in many navigation systems, especially the lidar-centric ones. However, they may no longer be effective with drastic changes in roll, pitch and vertical motions commonly seen in a UAV dataset. Besides, due to their long endurance, ground vehicle datasets can be conducted over a long period of time and a large environment, and have access to GPS. In these cases GPS data can be used for loop closure and/or ground truth, however it has significant error and reduces the reliability and comparability of the analysis. In contrast, the UAV datasets often have a shorter timespan and cover a smaller environment, which make them less representative of real-world scenarios.
However, for semi-controlled environment, UAV datasets can feature some artificial landmarks, i.e. visual markers or ranging anchors, to help reduce localization drift, which is applicable in most targeted industry inspection applications. Moreover, we can also employ high-accuracy laser tracking methods to provide ground truth, which is desirable for a stringent analysis and comparison of the localization methods.

\subsection{A new dataset for autonomous drone.}

\begin{figure}
    \centering
    \includegraphics[width=0.95\linewidth]{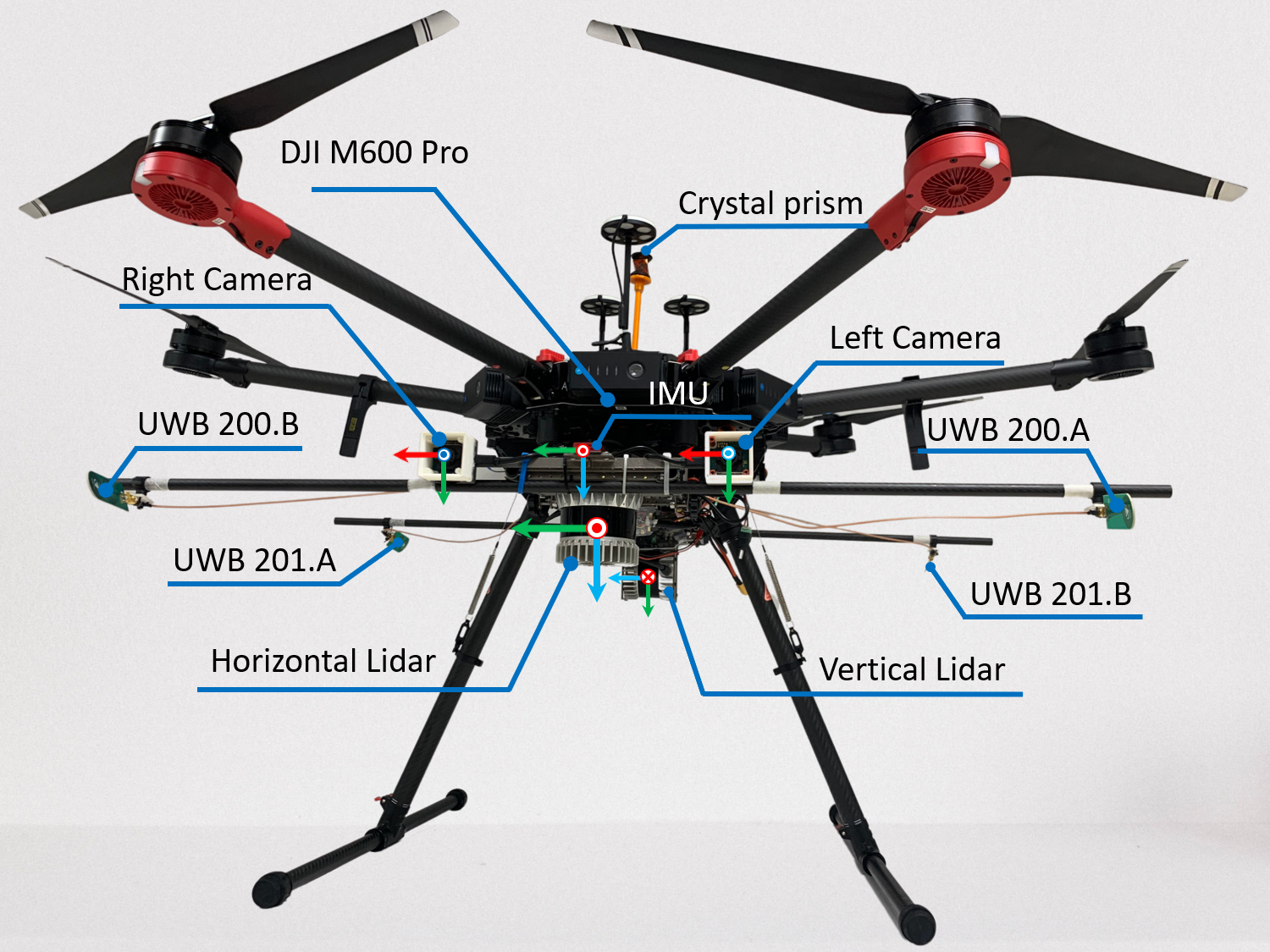}
	\caption{The drone setup used for the data collection. A DJI M600 Hexacopter is used to carry the payload comprising of 1 IMU, 2 lidars, 2 Cameras, 4 UWB ranging nodes, and a crystal prism to be tracked by a total station to provide ground truth.}
	\label{fig: hardware setup}
\end{figure}

Despite the aforementioned differences in the two classes of datasets, there is no doubt on their usefulness for the intended applications. As mentioned earlier, for the applications of aerial vehicles in our interest (Fig. \ref{fig: environments}), there appears to be an absence of compatible public dataset for the relevant scenarios.
This motivates us to construct a novel sensor setup and offer a new benchmark suite for advanced autonomous aerial systems that can be adopted by the industry in the near future.
Fig. \ref{fig: hardware setup} gives an overview of sensor setup.

Briefly speaking, we employ a high-frequecy IMU and a stereo camera rig typical of UAV systems (Sec. \ref{sec: rosbag imu}, Sec. \ref{sec: rosbag cameras}).
Next, thanks to the current advances in lidar hardware that have brought down its size and price, we find that it is due time 3D lidar can be popularly used on aerial platforms, thanks to many benefits of 3D lidar over a pure vision system.
Hence, we integrate two 3D lidars on our UAV platform. Different from previous datasets with 3D lidar mounted on ground vehicles, ours features a much more complex motion in 3D space with frequent drastic rotational and translational moves. More details on the lidars can be found in Sec. \ref{sec: rosbag lidars}.

In addition to the aforementioned sensors, as highlighted in Table \ref{tab: datasets and sensors}, our dataset also features UWB range measurements. Indeed, it is a consensus that drift is an intrinsic feature of onboard self-localization. However, our recent works have shown that by using range measurements to some fixed landmarks in an environment, one can reduce, or even eliminate such drift \cite{nguyen2020liro, nguyen2020tightlyauro}. For semi-controlled environments such landmarks can be easily deployed without much effort.
More information on this sensor is given in Sec. \ref{sec: rosbag uwbs}.

Finally, we employ a laser tracking system to provide high-accuracy\footnote{\url{https://github.com/ntu-aris/ntu_viral_dataset/blob/gh-pages/docs/Leica_Nova_MS60_DS.pdf}} ground truth for the position estimate. The experiments are conducted at various environments on the campus of Nanyang Technological University (NTU), including both indoor and outdoor conditions. The datasets are also recorded in rosbag format, which can be directly used on Robot Operation System (ROS). We also include some codes that can be used for the analysis of the localization information using our datasets. Several plug-and-play examples of using state-of-the-art localization methods with our datasets such as OpenVINS \cite{geneva2020openvins}, VINS-Fusion
\cite{qin2017vins},
A-LOAM
\cite{zhang2018laser},
LIO-SAM
\cite{shan2020liosam},
M-LOAM
\cite{jiao2020robust} are also provided. All of these resources can be found on our data suite's web page \url{https://ntu-aris.github.io/ntu_viral_dataset/}.

The rest of this paper is organized as follows: Sec. \ref{sec: sensor setup} presents our hardware setup and the intended purpose for each sensor. Sec. \ref{sec: dataset characteristics} presents the description of the environments and the flight tests. Sec. \ref{sec: dataset format} describes the organization, format, definitions and convention used in the datasets. Sec. \ref{sec: calibraion} explains the calibration procedures. Sec. \ref{sec: evaluation} describes our recommendation for evaluation. Sec. \ref{sec: issues} lists out the known issues in our implementation.

\section{Sensor setup} \label{sec: sensor setup}

A DJI M600 Pro\footnote{\url{https://www.dji.com/sg/matrice600-pro}} hexacopter is used to carry the sensor setup (Fig. \ref{fig: hardware setup}). Table \ref{tab: rosbag} provides a summary of the sensors and their corresponding specifications. All of the messages are timestamped by their \textit{publish time} on ROS. Below, we provide a more detailed description of these sensors.

\begin{table*}
\renewcommand{\arraystretch}{1.25}
\caption{The sensors used in this dataset and their corresponding specifications}
\label{tab: rosbag}
\resizebox{\linewidth}{!}
{%
\centering
\begin{tabular}{lllllc} 
\hline\hline
\textbf{No.}
& \textbf{Sensor} & \textbf{Model} & \textbf{Topic name} & \textbf{Message Type} &\textbf{Rate (Hz)}\\\hline

\mr{3}{*}{1}      & \mr{3}{*}{\begin{tabular}[c]{@{}l@{}}IMU\\(Sec. \ref{sec: rosbag imu})\end{tabular}} & \mr{3}{*}{\begin{tabular}[c]{@{}l@{}}VectorNav\\(VN100)\end{tabular}}
   & \texttt{/imu/imu}                  & sensor\_msgs/Imu           & 385\\
&& & \texttt{/imu/magnetic\_field}      & sensor\_msgs/MagneticField & 385\\
&& & \texttt{/imu/temperature}          & sensor\_msgs/Temperature   & 385\\\hline

\mr{2}{*}{2}    & \mr{2}{*}{\begin{tabular}[c]{@{}l@{}}Horizontal Lidar\\(Sec. \ref{sec: rosbag lidars})\end{tabular}} & \mr{2}{*}{OS1-16 Gen1}
   &\texttt{/os1\_cloud\_node1/imu}    & sensor\_msgs/Imu           & 100\\
&& &\texttt{/os1\_cloud\_node1/points} & sensor\_msgs/PointCloud2   & 10\\\hline

\mr{2}{*}{3}    & \mr{2}{*}{\begin{tabular}[c]{@{}l@{}}Vertical Lidar\\(Sec. \ref{sec: rosbag lidars})\end{tabular}} & \mr{2}{*}{OS1-16 Gen1}
    & \texttt{/os1\_cloud\_node2/imu}    & sensor\_msgs/Imu           & 100\\
&&  & \texttt{/os1\_cloud\_node2/points} & sensor\_msgs/PointCloud2   & 10\\\hline

4   & \begin{tabular}[c]{@{}l@{}}Camera 1\\(Sec. \ref{sec: rosbag cameras})\end{tabular} & {\begin{tabular}[c]{@{}l@{}}uEye 1221 LE\\(monochrome)\end{tabular}}
    & \texttt{/left/image\_raw}          & sensor\_msgs/Image         & 10\\\hline
    
5   & \begin{tabular}[c]{@{}l@{}}Camera 2\\(Sec. \ref{sec: rosbag cameras})\end{tabular} & {\begin{tabular}[c]{@{}l@{}}uEye 1221 LE\\(monochrome)\end{tabular}}
    & \texttt{/right/image\_raw}         & sensor\_msgs/Image         & 10\\\hline
    
\mr{3}{*}{6}
    & \mr{3}{*}{\begin{tabular}[c]{@{}l@{}}UWB Sensors\\(Sec. \ref{sec: rosbag uwbs})\end{tabular}} &\mr{3}{*}{Humatics P440}
    & \texttt{/uwb\_endorange\_info}     & uwb\_driver/UwbRange       & 68.571\\
&&  & \texttt{/uwb\_exorange\_info}      & uwb\_driver/UwbEcho        & 5.714\\
&&  & \texttt{/nodes\_pos\_sc}           & nav\_msgs/Path             & 5.714\\\hline

7   & \begin{tabular}[c]{@{}l@{}}3D Laser Tracker\\(Sec. \ref{sec: rosbag leica})\end{tabular} & \begin{tabular}[c]{@{}l@{}}MS60 Leica\\TotalStation\end{tabular}
    & \texttt{/leica/pose/relative}      & \begin{tabular}[c]{@{}l@{}}geometry\_msgs/PoseStamped\\(orientation is not set)\end{tabular} & 20\\
\hline\hline
\end{tabular}
}
\end{table*}

\subsection{IMU:} \label{sec: rosbag imu}
A VectorNav VN100\footnote{\url{https://www.vectornav.com/products/VN-100}} rugged IMU is employed as the main inertial sensor in our system. It is also chosen to be the center of the \textit{body frame} where the extrinsics of other sensors are referenced to. The sensor is configured to publish data at 400Hz, though the effective rate is roughly 385Hz. Note that besides the angular velocity and acceleration measurements, the topic \texttt{/imu/imu} also contains the orientation estimate by the device's internal Extended Kalman Filter, which also fuses magnetometer with the aforementioned measurements.

\subsection{3D lidars:} \label{sec: rosbag lidars}
In this work, two 16-channel OS1 gen1\footnote{\url{https://ouster.com/products/os1-lidar-sensor}} lidars are configured so that the so-called \textit{horizontal lidar} can scan the objects in the front, back, left, and right sides of the UAV, while the other so-called \textit{vertical lidar} can scan the ground, front and back sides. Thus, the two can complement each other well to maintain good observation on the environment. Note that each lidar has an internal IMU that outputs angular rate and acceleration measurements at 100Hz rate. Each lidar is loaded with the latest V2 firmware for additional fields such as reflectivity, time, and ambient. These quantities are also recorded in the bag files.

\subsection{Stereo cameras} \label{sec: rosbag cameras}
Two uEye 1221 LE\footnote{\url{https://en.ids-imaging.com/store/ui-1221le-rev-2.html}} monochrome-global-shutter cameras are mounted on the drone, facing directly forward. The two are synchronized by an external trigger to capture and publish image at almost the same time. \textit{Typically, the difference in the timestamps of the images triggered at the same time is below 3 ms}. This is necessary if one needs to set a threshold to synchronize the images in the buffer for stereo processing such as in VINS-Fusion \cite{qin2017vins}.

\subsection{UWB ranging sensors} \label{sec: rosbag uwbs}
The UWB ranging network is setup in a similar manner with our previous works \cite{nguyen2020liro, nguyen2021viral, nguyen2018robust}. Two Humatics P440 UWB radios\footnote{\url{https://humatics.com}}, given the IDs 200 and 201 in the network, are mounted on the UAV, each has two antennae A and B extended to the four corners. Hence we have a total of 4 ranging nodes on the UAV: 200.A, 200.B, 201.A, 201.B (see Fig. \ref{fig: hardware setup}). 
Another three UWB nodes with IDs 100, 101, 102 are used as anchors. As such in total we have 12 UAV-to-anchor ranging pairs.

\begin{figure}
    \setlength\belowcaptionskip{-0.25cm}
    \centering
    \includegraphics[width=\linewidth]{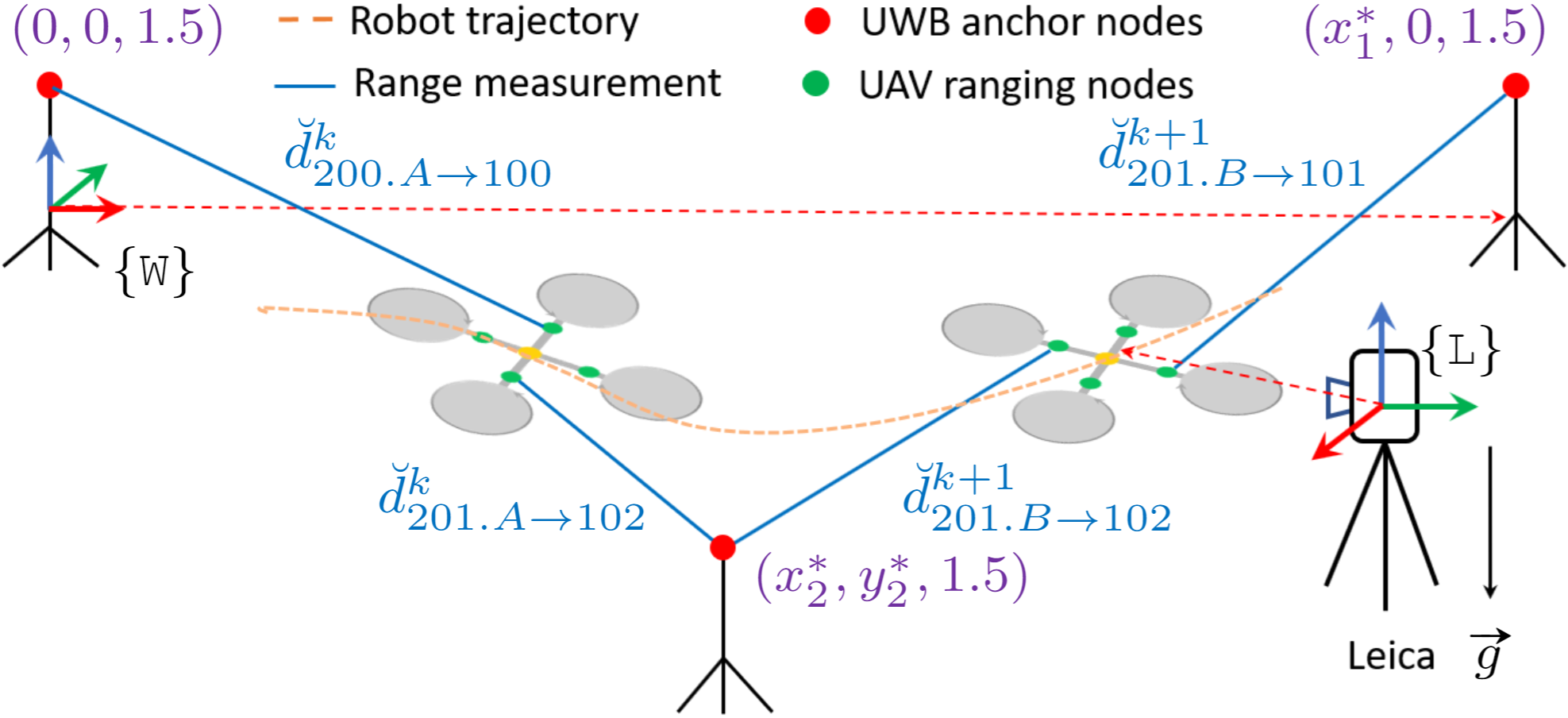}
	\caption{Illustration of the ranging scheme and the external frame of references. Refer to Sec. \ref{sec: rosbag uwbs} for detailed descriptions.
	}
	\label{fig: ranging scheme}
\end{figure}

The ranging sequence is programmed in a way such that all 12 ranging pairs are executed equally.
Each ranging step takes 25 ms. Hence with only one UAV ranging node, ideally one can expect 40 Hz of ranging measurements on the topic \texttt{/uwb\_endorange\_info}. However, in our case, there are two ranging nodes (we program node 200 to range to one anchor while 201 ranges to another), thus we can obtain twice the amount of range measurements, i.e. 80 Hz. Nevertheless, we reserve some steps in the sequence to let the anchors range among themselves and broadcast the measurements. Therefore the rate of the UAV-to-anchor ranges, i.e. the \texttt{/uwb\_endorange\_info} topic is reduced to 68.571 Hz, while the rate of anchor-to-anchor ranges, published over the \texttt{/uwb\_exorange\_info} topic, is 5.714 Hz, as shown in Tab. \ref{tab: rosbag}. Fig. \ref{fig: ranging scheme} gives a simple illustration of the ranging incidents over two time steps. Note that due to the fact that the transmission power of the UWB depends on the relative orientation between the antennae of the the two nodes, some ranging steps may fail to return, or return an unreliable measurements. Thus the effective rate of UWB measurement can vary through time, depending on the position of the UAV relative to the anchors during flight.

The anchors are placed on tripods and they are adjusted to have the same height from the ground. Thus, their nominal z coordinate is 1.5m. If we select anchor 100 to be at the origin, and anchor 101 to be on the +x direction, and +z direction is from the ground to the anchor nodes, then we can establish a coordinate system using the right hand rule (the frame $\{\fW\}$ in Fig. \ref{fig: ranging scheme}). Using this arrangement and the distances from the topic \texttt{/uwb\_exorange\_info} we can easily estimate the coordinates $x_1^*$, $x_2^*$, $y_2^*$. The coordinates of the anchor nodes are published in the \texttt{/nodes\_pos\_sc} topic.

\subsection{Ground truth} \label{sec: rosbag leica}
A Leica Nova MS60 MultiStation\footnote{\url{https://Leica-geosystems.com/en-sg/products/total-stations/multistation/Leica-nova-ms60}} is used to track a crystal prism mounted on the top of the UAV to provide ground truth for position estimate. Note that the coordinate frame of this ground truth system is aligned with gravity during the startup, and its z axis points at the opposite direction of the gravity (Fig. \ref{fig: ranging scheme}).
Moreover, since there is a significant displacement between the prism and the body frame's origin, the accuracy analysis needs to take this into account. Sec. \ref{sec: evaluation} discusses this issue in details.

\section{Dataset characteristics} \label{sec: dataset characteristics}

\begin{figure*}
	\centering
	 \begin{subfigure}[h]{0.32\linewidth}
        \centering
        \includegraphics[width=\linewidth]{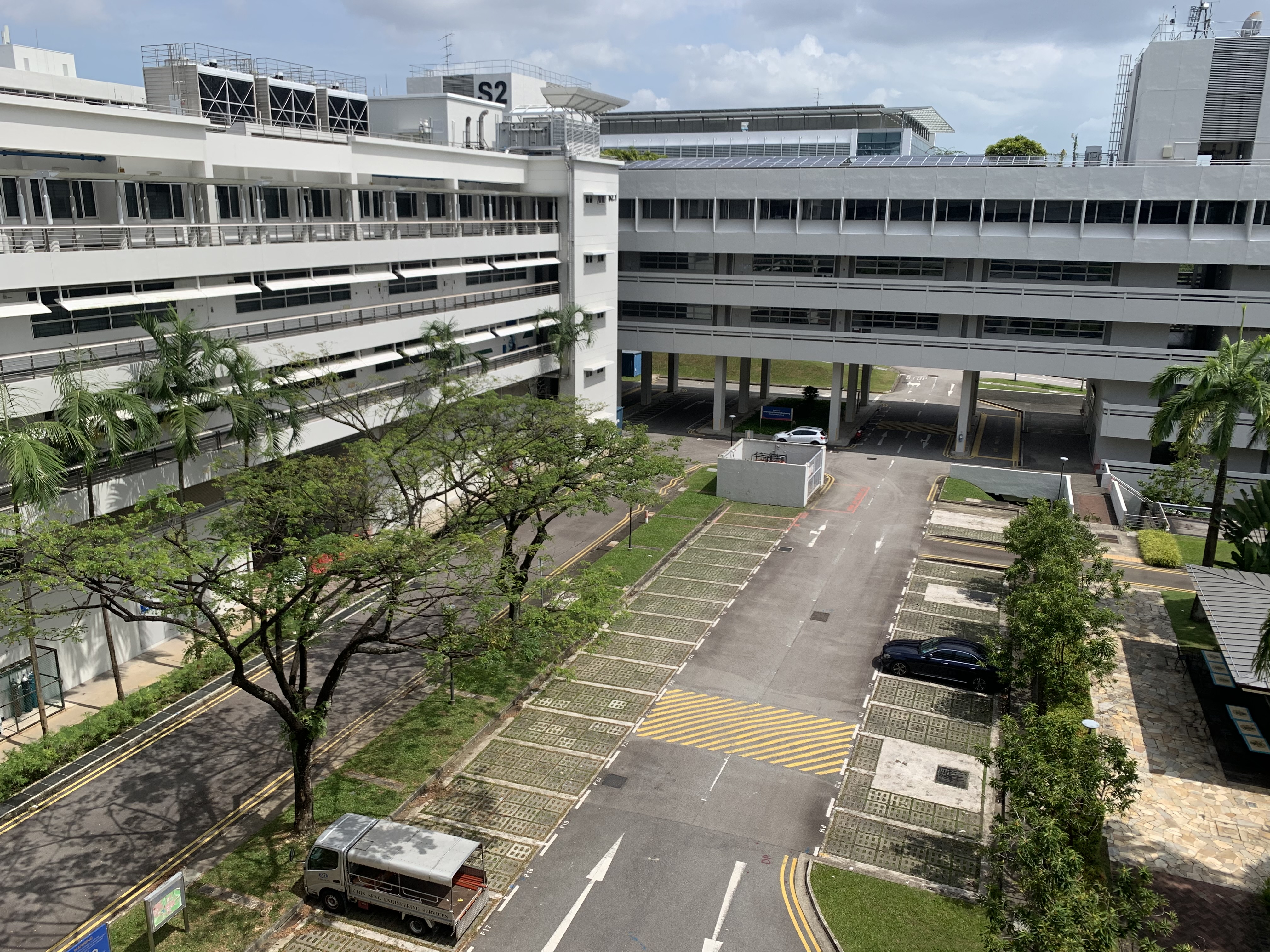}
		\caption{EEE environment}
		\label{fig: eee environtments}
	\end{subfigure}
	\hfill
    \begin{subfigure}[h]{0.32\linewidth}
		\includegraphics[width=\linewidth]{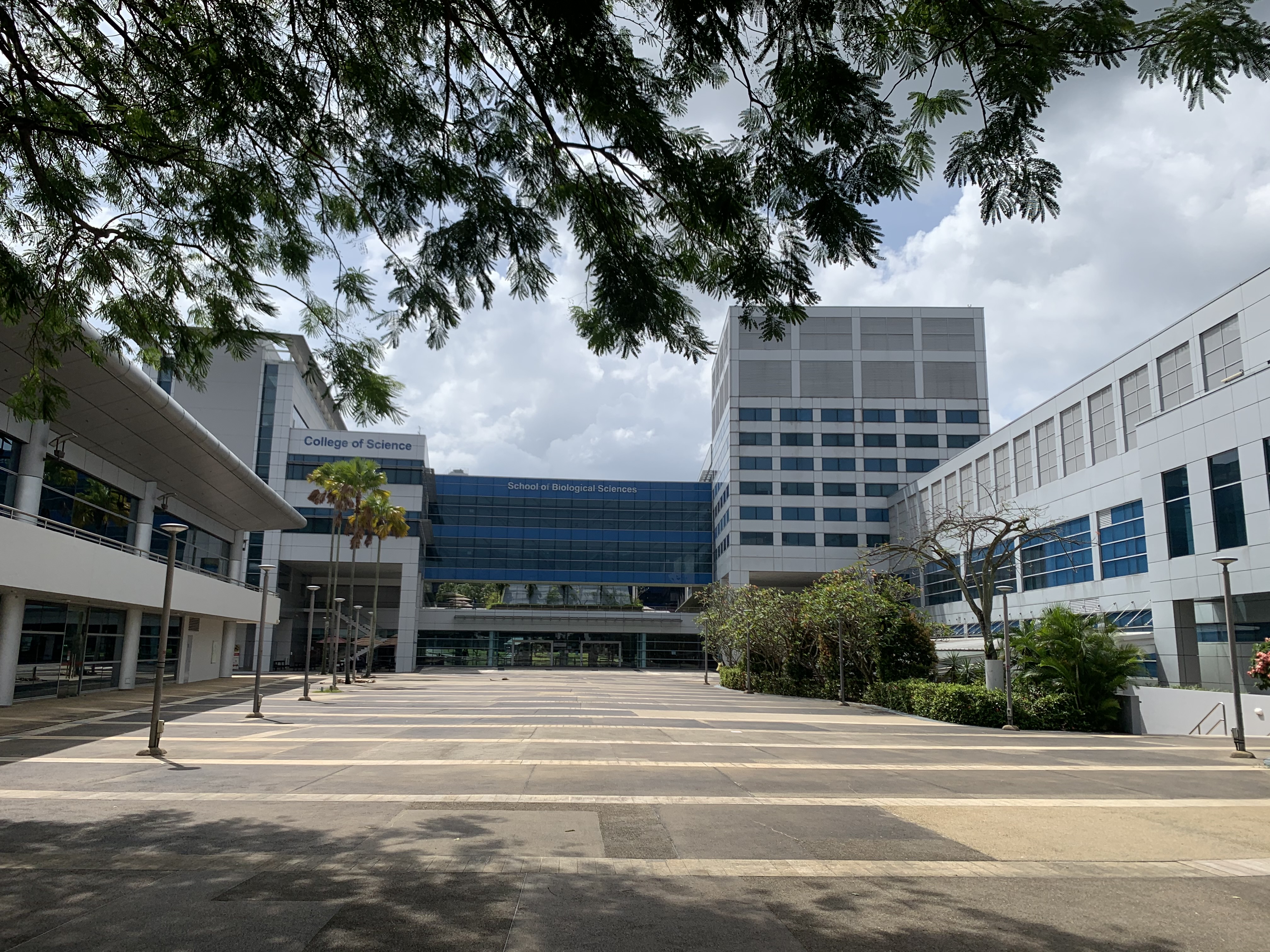}
		\caption{SBS environment}
		\label{fig: sbs environments}
	\end{subfigure}
	\hfill
    \begin{subfigure}[h]{0.32\linewidth}
		\includegraphics[width=\linewidth]{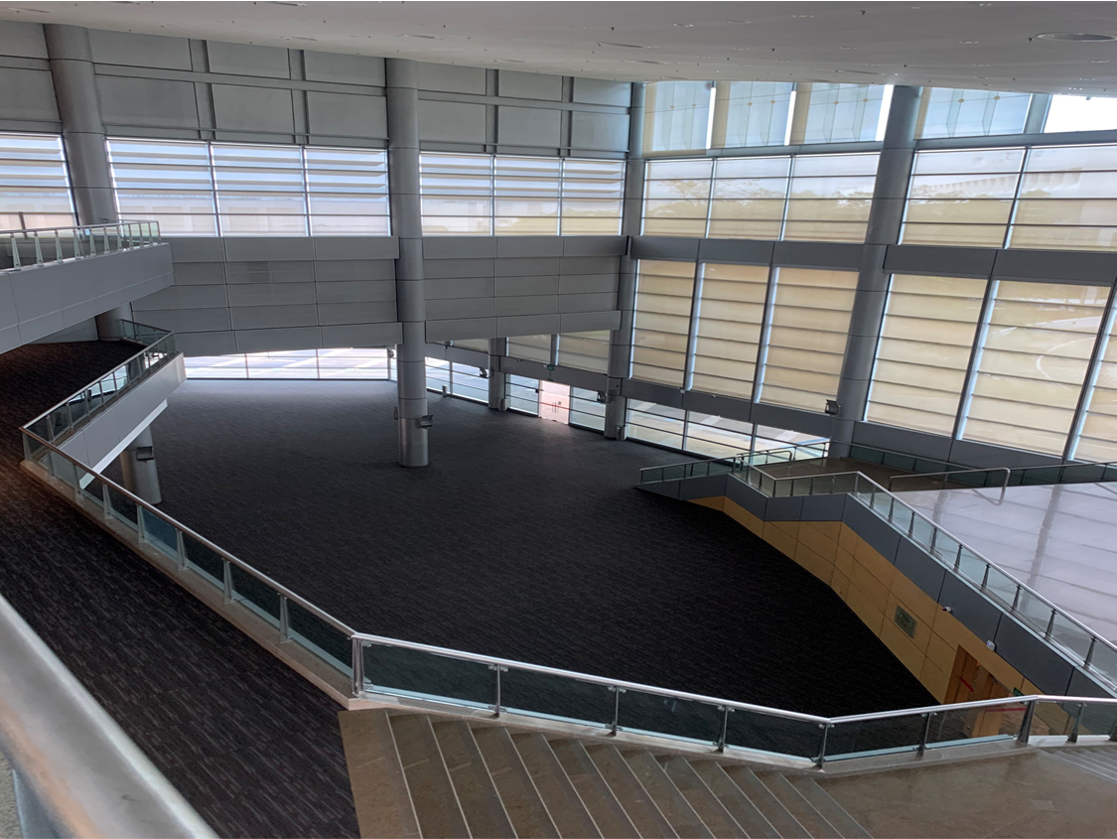}
		\caption{NYA environment}
		\label{fig: nya environments}
	\end{subfigure}
	\caption{Environments where the datasets are collected}  \label{fig: ntu environments}
\end{figure*}

The datasets are divided into three groups based on the environments, namely the EEE, SBS and NYA sequences. The environments include both indoor and outdoor locations on NTU campus. Fig. \ref{fig: ntu environments} presents some overviews of these environments.
Tab. \ref{tab: dataset features} gives a brief summary of the sequences and their statistics.

\begin{table}
\caption{Some statistics of each sequence} \label{tab: dataset features}
\begin{tabular}{ccccc}
\hline
\textbf{Name} & \textbf{Time} & \textbf{\begin{tabular}[c]{@{}c@{}}Path\\ Length\end{tabular}} & \textbf{\begin{tabular}[c]{@{}c@{}}Average\\ Vel.\end{tabular}} & \textbf{\begin{tabular}[c]{@{}c@{}}Average\\ Ang. Vel\end{tabular}} \\ \hline
eee\_01     & 398.7 s   & 237.07 m  & 0.677 m/s     & 0.119 rad/s   \\
eee\_02     & 321.1 s   & 171.07 m  & 0.585 m/s     & 0.101 rad/s   \\
eee\_03     & 181.4 s   & 127.83 m  & 0.800 m/s     & 0.184 rad/s   \\
nya\_01     & 396.3 s   & 160.30 m  & 0.463 m/s     & 0.150 rad/s   \\
nya\_02     & 428.7 s   & 249.10 m  & 0.657 m/s     & 0.141 rad/s   \\
nya\_03     & 411.2 s   & 315.46 m  & 0.821 m/s     & 0.125 rad/s   \\
sbs\_01     & 354.2 s   & 202.30 m  & 0.625 m/s     & 0.101 rad/s   \\
sbs\_02     & 373.3 s   & 183.57 m  & 0.573 m/s     & 0.145 rad/s   \\
sbs\_03     & 389.3 s   & 198.54 m  & 0.601 m/s     & 0.120 rad/s   \\ \hline
\end{tabular}
\end{table}

\subsection{The EEE sequences}
Three datasets are collected at the carpark in the center of the School of EEE, NTU, hence named the \textit{EEE sequences}. The area is surrounded by tall building blocks on all side which can be conducive for lidar-based SLAM. Also, reliable visual features can be detected on nearby objects such as trees, road markings, and buildings (see Fig. \ref{fig: eee environtments}).


\subsection{The SBS sequences}
\begin{figure}[h]
    \centering
    \includegraphics[width=\linewidth]{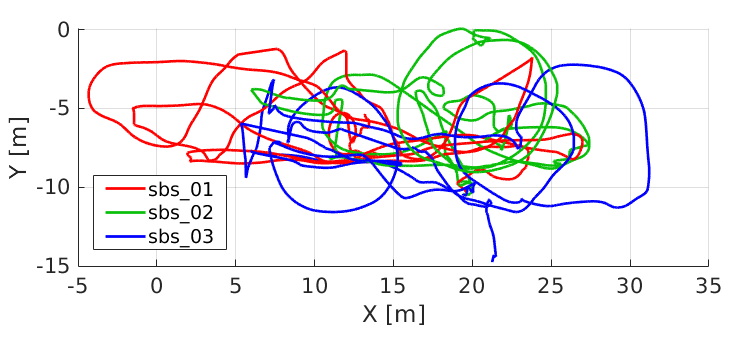}
	\caption{Trajectories of the AV in the SBS sequences as recorded by the Leica laser tracker.}
	\label{fig: SBS sequences}
\end{figure}

The \textit{SBS sequences} are collected at an open square next to the School of Biological Sciences, NTU. This area is surrounded by some low-rise buildings with large glass surfaces. Also, visual features may only be detected on objects far way, which can produce noisy depth.
The drone's trajectories in these sequences are shown in Fig. \ref{fig: SBS sequences}.

\subsection{The NYA sequences}
\begin{figure*}
	\centering
	 \begin{subfigure}[h]{0.495\linewidth}
        \centering
        \includegraphics[width=\linewidth]{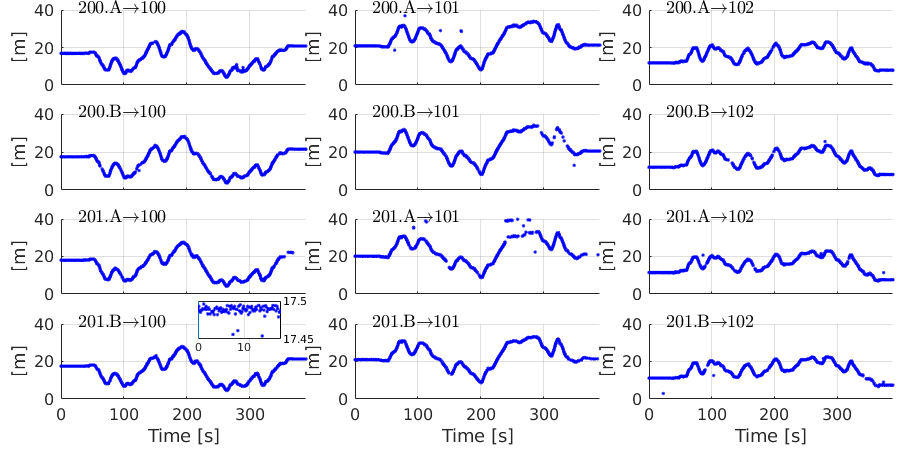}
		\caption{UWB ranges in sbs\_03 dataset. The zoom-in plot shows a 5 cm fluctuation in measurement when the drone was static.}
		\label{fig: uwb sbs 03}
	\end{subfigure}
	\hfill
    \begin{subfigure}[h]{0.495\linewidth}
		\includegraphics[width=\linewidth]{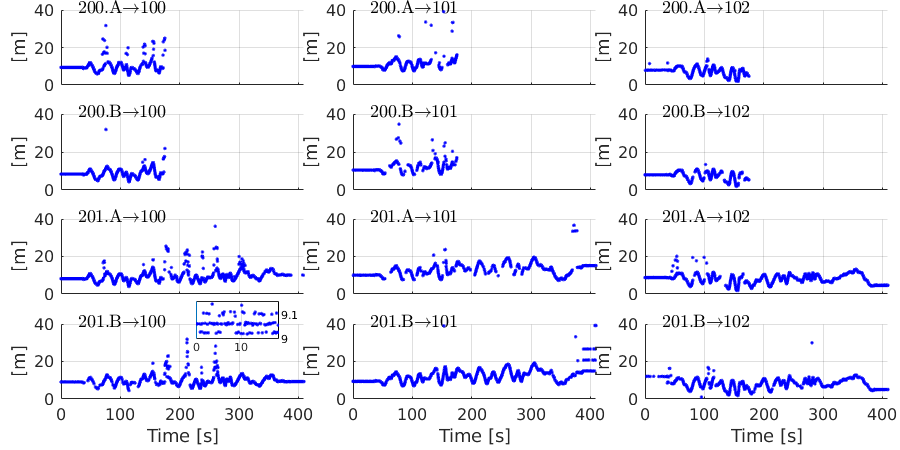}
		\caption{UWB ranges in nya\_03 dataset, showing signal loss by UWB 200 mid-flight, and multi-path effect on all ranging edges}
		\label{fig: uwb nya 03}
	\end{subfigure}
	\caption{UWB ranges in outdoor and indoor conditions}  \label{fig: uwb comparison}
\end{figure*}

The auditorium at NTU, shown in Figure \ref{fig: nya environments}, represents a challenging environment for visual SLAM. These datasets feature some significant challenges. For e.g. semi-transparent surfaces can be an issue for lidar SLAM, while flight dynamics and low lighting conditions are difficult for visual SLAM. Moreover, we also notice significant multi-path effects and signal loss on the UWB measurements (Fig. \ref{fig: uwb comparison}).


\section{Dataset format}  \label{sec: dataset format}

\subsection{Files organization}
The datasets are recorded into rosbag files during the flight, and saved on the memory of the drone's onboard computer (DJI Manifold 2C). The ground truth data is recorded on a seperate computer, and is later temporally synchronized and merged with the bag from the drone's onboard computer, yielding a single rosbag for each flight test.

\begin{figure}[h]
    \centering
    \includegraphics[width=\linewidth]{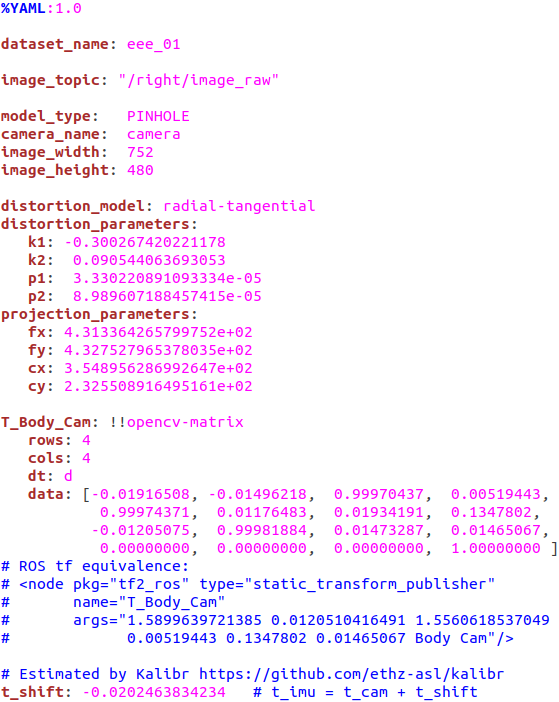}
	\caption{Content of the calibration file camera\_right.yaml of the eee\_01 dataset.}
	\label{fig: camera calib file}
\end{figure}

Each rosbag file is accompanied by a set of files containing the calibration parameters for each sensor. Fig. \ref{fig: camera calib file} illustrates the content of such a calibration file. These files document the coefficients of the camera model, sensor-to-body coordinate transforms, IMU noise, camera-IMU time delay, and can be parsed using opencv and ROS APIs. In addition, some accessories to utilize the pointclouds and UWB data are also provided with links on the data suite's web page.

\subsection{Units and timestamps}
All of the measurements are reported in SI units except the timestamps of the messages are in \texttt{ros/Time}\footnote{\url{http://wiki.ros.org/roscpp/Overview/Time}} format, which can be found in the \texttt{header.stamp} field of the messages. The timestamps are assigned by each sensor's driver when the message is published over ROS.

In addition, the timestamp of each lidar pointcloud message matches with the \textit{the end-time of the scan}. Moreover, the pointcloud also contains the timestamp of each point relative to \textit{the start-time of the scan}. This will be useful if one seeks to perform deskew operation on the pointcloud.

\subsection{Message types}

The message types for most sensors are standard ROS messages, however there are some custom definitions for UWB. Specifically, for the UWB messages of type \texttt{uwb\_driver/UwbRange}, besides the distance measurement, the message also contains other information such as signal over noise ratio, line-of-sight self-diagnosis, antenna coordinates in the body frame, and the anchor coordinates, which are fixed by using the first messages in \texttt{/nodes\_pos\_sc}. The source code that defines these message can be found on our data suite's web page \url{https://ntu-aris.github.io/ntu_viral_dataset}.

\subsection{Coordinate transformations}

For common objects such as vectors, quaternions, rotation matrices recorded in the rosbag files, the ROS conventions are followed. Here we shall explain the extrinsics defined in the calibration files.

Let us take an instance to explain the convention used throughout the datasets. In reference to Fig. \ref{fig: camera calib file}, notice that the transform ${}^{\fB}_\fC\tf \in SE(3) \subset \mathbb{R}^{4\times4}$ is declared via the parameter \texttt{T\_Body\_Cam} (the "body-cam" suffix follows a parent-child order between the two frames of reference, following the convention of ROS and the EuRoC dataset). This transform consists of the rotation ${}^{\fB}_\fC\rot \in SO(3) \subset \mathbb{R}^{3\times3}$ on the upper left corner and the translation ${}^{\fB}_\fC\trans \in \mathbb{R}^{3\times1}$ on the upper right corner, i.e. ${}^{\fB}_\fC\tf = \begin{bmatrix}{}^{\fB}_\fC\rot &{}^{\fB}_\fC\trans\\0 &1\end{bmatrix}$.


For an object $\fA$ observed in the camera frame $\{\fC\}$ with position ${}^{\fC}_\fA\pos$ and orientation ${}^{\fC}_\fA\rot$, its corresponding position and orientation in the body frame ${\fB}$, denoted as ${}^{\fB}_\fA\pos$ and ${}^{\fB}_\fA\rot$, can be obtained by:
\begin{equation} \label{eq: coordinate conversion}
    {}^{\fB}_\fA\pos = {}^{\fB}_\fC \rot {}^{\fC}_\fA\pos + {}^{\fB}_\fC\trans,\ \ 
    {}^{\fB}_\fA\rot = {}^{\fB}_\fC \rot {}^{\fC}_\fA\rot.
\end{equation}
The object $\fA$ in this case can be a 3D feature or an object that is being tracked by the camera or the lidar, which in a typical SLAM algorithm needs to be coupled with the extrinsics in the estimation process. Moreover, $\fA$ can be another coordinate system that we need to convert the coordinates from other frames of references to.


\section{Sensor calibration} \label{sec: calibraion}

Sensor fusion often requires prior knowledge of the spatial configuration of multiple sensors. This section introduces our calibration pipelines, which is divided into two parts, namely the rigid systems and flexible systems. The cameras and IMU are mounted on the same rigid titanium-alloy-based 3D-printed parts. Thus, their spatial relation has very small dynamic variance. The other sensors are flexibly mounted as they are connected via some carbon fiber tubes or dampers. These flexible parts will have large dynamic variance.

We first calibrate the rigid body system. The stereo cameras are calibrated both intrinsically and extrinsically. We then find the IMU to camera transform based on visual-inertial alignment. For the more flexible parts, the transform may come with more considerable variance in the spatial relation. Therefore, we mostly use the VICON system to calibrate between those sensors with some help from other priors such as recognizable planes in the lab. All of the calibration results can be verified using localization or mapping packages.

\subsection{Stereo calibration}
Our first goal is to find the intrinsic and extrinsic parameters of the stereo cameras using an approach similar to \cite{zhang2000flexible}. We collect N well-synchronized stereo images in front of a static calibration chessboard to minimize the motion blur. Using pinhole camera and radial-tangential distortion models, the intrinsics are calibrated using all of the image sequences on the MATLAB calibration toolbox. We set a threshold for the reprojection error and only use the smaller subset of images for the extrinsics. We find that a large roll of chessboard pattern will enlarge the reprojection error for this 120-degree FOV lens. Therefore, we keep a small roll factor in calibration sequences. Finally, we verify the calibration result by comparing the stereo-matching reprojected depth to lidar measurement. The result can be well aligned with the lidar. The stereo calibration image sequence is provided in the dataset portal.

\subsection{Visual inertial calibration}
We calibrate and refine the intrinsic and extrinsic parameters of IMU and cameras by using the camera-IMU calibration procedure\footnote{\url{https://github.com/ethz-asl/kalibr/wiki/camera-imu-calibration}} of the Kalibr package \cite{furgale2013unified}. In this procedure, the initial guess of the camera intrinsic parameters is required and we obtain them from the stereo calibration process in the previous part. The intrinsics (biases) of IMU are calibrated beforehand using the manufacturer's software\footnote{\url{https://www.vectornav.com/resources/software}}, and the correction is applied to the raw inertial measurement. At last, we collect the visual-inertial sequence with full 6 DOF excitation and perform the visual-inertial calibration. We have test the calibration result using VINS-Mono, VINS-Fusion, OpenVINS, and we can obtain meaningful trajectories using the calibrated parameters. A 20-ms time offset between IMU and camera timestamps is found by the Kalibr package, which is also reported in the accompanying yaml file (Fig. 9). However this value is just for reference, and we do not make any modification on sensor's timestamps. The rosbag files used for calibration can also be found at the website.

\subsection{Other flexible parts compensation and ground truth}
The dataset is captured from a payload bay mounted on our custom DJI M600 Pro drone. Therefore, the triple GPS, compass, and prism on top of the drone are not rigidly connected to the payload bay, i.e. their spatial relation can change during flight. However, this is a necessary compromise to reduce vibration effect on the quality of the lidar scan, camera image, and IMU.
In light of these conditions, we first use VICON to verify most of the other extrinsic parameters by putting the VICON reflective ball at each sensor while the drone is at rest on the floor.
Then we exert some force on the dampened payload bay to tilt it up to some maximum angle. The maximum displacement of the prism between static configuration and maximum tilting angle is around 2cm. Based on our study, the drone only exhibits small control angles with a minimized jerk during the flight tests. Therefore, the prism measurement is relatively accurate to represent the ground truth. Note that we use hand-eye calibration to align the ground truth with onboard altitude estimate obtained from the drone’s internal barometer and IMU, which are independent from the sensors on the payload.

\begin{figure*}
    \centering
    \includegraphics[width=\linewidth]{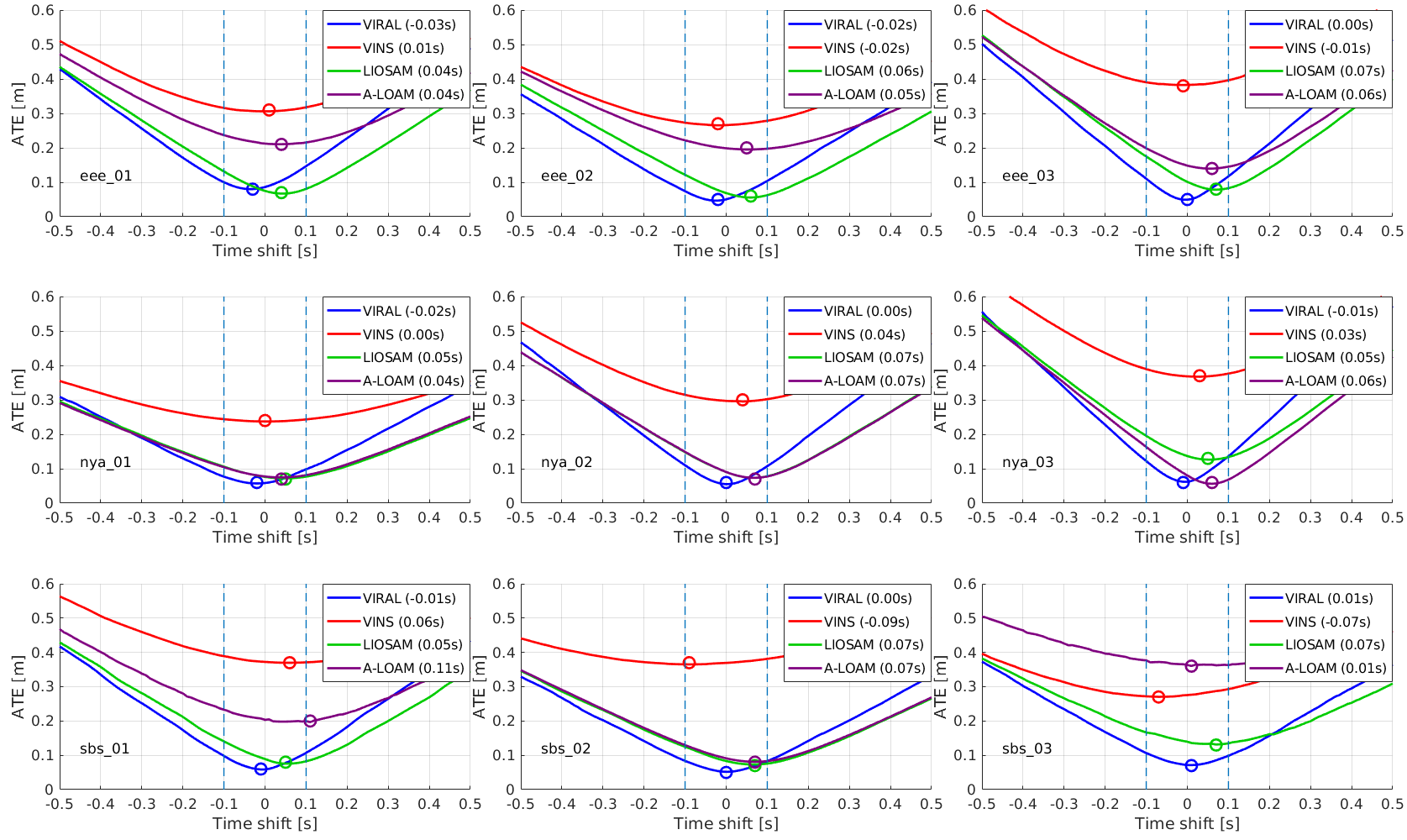}
	\caption{\highlight{The changes in ATE of different methods with respect to time shifts of the ground truth. Note that except A-LOAM the selected methods all feature loop closure and global optimization to eliminate the influence of drift. The minimum of each curve is marked by a circle, and the time shift that minimizes the ATE for each method is reported in the legend. It can be seen that all visual-inertial, lidar-inertial, and visual-inertial-lidar-ranging methods can achieve minimum error within the [-0.1, 0.1] interval.}}
	\label{fig: ATE vs time shift}
\end{figure*}

\begin{figure}
    \centering
    \includegraphics[width=\linewidth]{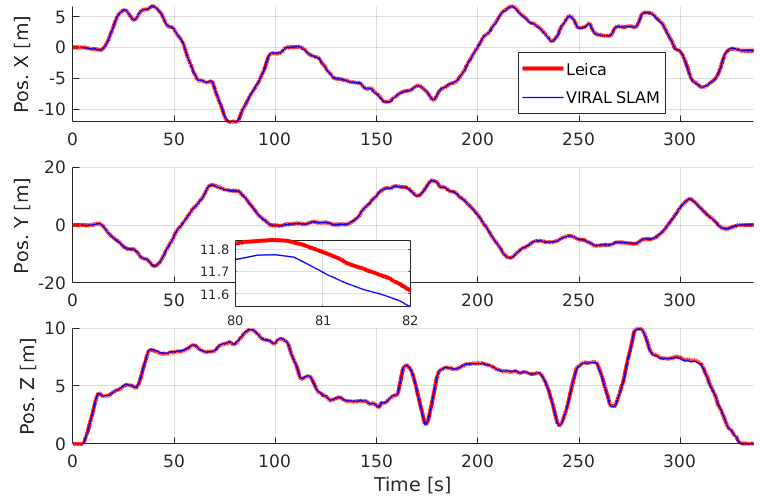}
    \hfill
    \includegraphics[width=\linewidth]{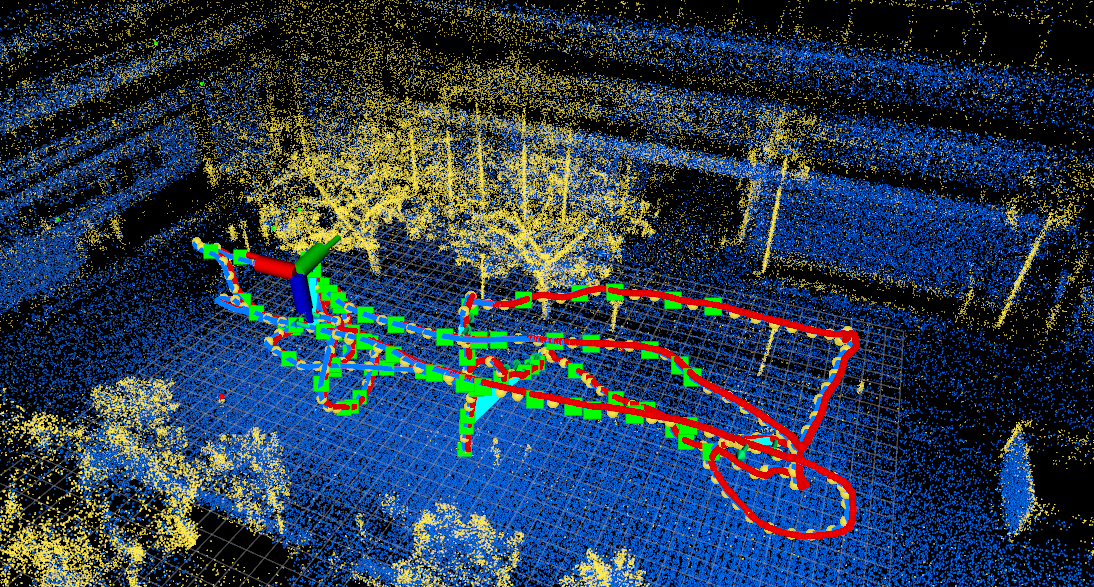}
	\caption{VIRAL SLAM result using the eee\_01 dataset. Ground truth is maked in red, the trajectory estimate in blue. The 3D map of the environment is visualized in blue and yellow points in the 3D visualisation below. Key frames are highlighted by green boxes. The ground truth and trajectory estimate are aligned using the 6DoF transformation calculated in \eqref{eq: coordinate alignment}.}
	\label{fig: demo}
\end{figure}

\highlight{To obtain a bound of the error in the temporal alignment between ground truth and onboard sensors, we conduct several experiments with several state-of-the-art localization methods and calculate their Absolute Trajectory Error (ATE) with multiple versions of ground truth that are time-shifted by -0.5, -0.49, -0.48,… 0.5 seconds. The results are reported in Fig. \ref{fig: ATE vs time shift}. We can see that despite using different sensor combinations and methods, i.e. stereo camera-IMU (VINS), lidar-IMU (LIO-SAM), and stereo camera-IMU-lidar-UWB (VIRAL SLAM), all methods can achieve minimal ATE within the [-0.1, 0.1] interval. Thus, we are confident that the temporal alignment error between ground truth and the onboard sensors is within 0.1s in the current datasets. Table \ref{tab: all ATE} presents the ATE of the SLAM frameworks without time shift. A more comprehensive analysis can be found in \cite{nguyen2021viralslam}. Fig. \ref{fig: demo} demonstrates a result of VIRAL SLAM on one sequence.}

\begin{table}
\centering
\caption{\highlight{ATE of state-of-the-art localization methods over NTU VIRAL datasets. The best odometry result is highlighted in \tb{bold}, and the second best is \ul{underlined}. All values are in meter.}} \label{tab: all ATE}
\renewcommand{\arraystretch}{1.1}
\begin{tabular}{cccccc}
\hline\hline
\tb{Dataset}
&\tb{\begin{tabular}[c]{@{}c@{}}VINS-\\Fusion\end{tabular}}
&\tb{\begin{tabular}[c]{@{}c@{}}LIO-\\SAM\end{tabular}}
&\tb{\begin{tabular}[c]{@{}c@{}}MLOAM\end{tabular}}
&\tb{\begin{tabular}[c]{@{}c@{}}VIRAL-\\SLAM\end{tabular}}\\ \hline
{eee\_01}
        &{0.608}                    
        &\ul{0.075} & {0.249}       
        &\tb{0.060} \\	            
{eee\_02}
        &{0.506}                    
        &\ul{0.069} & {0.166}       
        &\tb{0.058} \\	            
{eee\_03}
        &{0.494}                    
        &\ul{0.101} & {0.232}       
        &\tb{0.037} \\	            
{nya\_01}
        &{0.397}                    
        &\ul{0.076} & {0.123}       
        &\tb{0.051} \\	            
{nya\_02}
        &{0.424}                    
        &\ul{0.090} & {0.191}       
        &\tb{0.043} \\	            
{nya\_03}
        &{0.787}                    
        &\ul{0.137} & {0.226}       
        &\tb{0.032} \\	            
{sbs\_01}
        &{0.508}                    
        &\ul{0.089} & {0.173}       
        &\tb{0.048} \\	            
{sbs\_02}
        &{0.564}                    
        &\ul{0.083} & {0.147}       
        &\tb{0.062} \\	            
{sbs\_03}
        &{0.878}                    
        &\ul{0.140} & {0.153}       
        &\tb{0.054} \\	            
\hline\hline
\end{tabular}
\end{table}

\section{Evaluation recommendations} \label{sec: evaluation}

From our experiments of state-of-the-art SLAM methods on NTU VIRAL, we believe some issues regarding the evaluation process merit a detailed discussion.
In most cases, after applying a navigation method on our dataset, the user can obtain a trajectory consisting of $N+1$ position estimates of the body frame, denoted as $\{{}_{\fB}^\fW\hat{\pos}_{t_n}\}_{n = 0}^{N}$, and a trajectory of $M$ ground truth samples of the crystal prism $\{{}_\fP^{\fL}{\pos}_{t_m}\}_{m = 0}^{M}$. From these samples, the user would like to calculate the estimation accuracy of the method based on some metrics.
To facilitate the use of our dataset, MATLAB scripts for this purpose are also included in the suite. We note that some previous works such as \cite{sturm2012benchmark, zhang2018tutorial} might have briefly addressed some of the issues discussed below. However, we think a more formal description can be of interest to some readers.

\subsection{Are we tracking the same point?}
Since the crystal prism has a displacement of almost 0.4 m from the body frame's origin, if we naively calculate the error between the body-centered trajectory with the ground truth, there can be a bias of 0.4 m in the estimate in the worst case scenario.
Thus it is recommended that the user use the pose estimate to calculate the estimated trajectory of the crystal prism, denoted as $\{{}_{\fP}^\fW\hat{\pos}_{t_n}\}_{n = 0}^{N}$, for comparison with ground truth. Specifically, using \eqref{eq: coordinate conversion}, we can obtain:
\begin{equation}
    {}_{\fP}^\fW\hat{\pos}_{t_n} = {}_{\fB}^\fW\hat{\pos}_{t_n} + {}_{\fB}^\fW\hat{\rot}_{t_n} {}_{\fP}^\fB\trans_{t_n},\ \forall n \in \{0, 1, \dots N\},
\end{equation}
where ${}_{\fB}^\fW\hat{\rot}_{t_n}$ is the orientation estimate, and ${}_{\fP}^\fB\trans_{t_n}$ is the translation from the body origin to the prism reported in the dataset's calibration files.
Hence, we can drop the subscript ${}_{\fP}$ as it is implied in later parts.

\subsection{Resampling} \label{sec: resampling}

The goal of this task is to resample $\{{}^\fW\hat{\pos}_{t_n}\}_{n = 0}^{N}$ and $\{{}^\fL{\pos}_{t_m}\}_{m = 0}^{M}$ to obtain the sequences $\{{}^\fW\hat{\pos}_{t_k}\}_{k = 0}^{K}$ and $\{{}^\fL{\pos}_{t_k}\}_{k = 0}^{K}$ of the same length.
Specifically $\{t_k\}_{k=0}^K$ is a subset of $\{t_n\}_{n=0}^N$ that satisfies the following
\begin{align*}
  \exists\ t_s \in \{t_m\}_{m=1}^M \colon t_s \leq t_k \leq t_{s+1}, \vert t_s - t_{s+1}\vert < 0.1, \forall t_k.
\end{align*}
Given the time $t_s$ above, the ground truth sample at time $t_k$ can be linearly interpolated as ${}^{\fL}\pos_{t_k} = \frac{t_{k} - t_s}{t_{s+1} - t_s}{}^{\fL}\pos_{t_s} + \frac{t_{s+1} - t_k}{t_{s+1} - t_s}{}^{\fL}\pos_{t_{s+1}}$.
We can further simplify the notation by denoting the trajectories as $\{{}^\fW\hat{\pos}_{k}\}_{k=0}^{K}$ and $\{{}^\fL{\pos}_{k}\}_{k=0}^{K}$.

\subsection{Coordinates transform}
At the final stage, we can see that $\{{}^\fW\hat{\pos}_{k}\}_{k=0}^{K}$ and $\{{}^\fL{\pos}_{k}\}_{k=0}^{K}$ are still \wrt different coordinate frames. Hence they cannot be directly compared. Following the common public benchmarking tools by \cite{sturm2012benchmark} and \cite{zhang2018tutorial}, we adopt the following coordinate transform $({}^\fL_\fW\rot, {}^\fL_\fW\trans)$ to bring the two trajectories to a common frame of reference:
\begin{equation} \label{eq: coordinate alignment}
    ({}^\fL_\fW\rot,\ {}^\fL_\fW\trans) = \argmin_{(\rot,\ \trans)}\left(\sum_{k=0}^K\norm{\rot{}^\fW\hat{\pos}_{k} + \trans - \pos_{k}}^2\right).
\end{equation}
The above was shown to have a closed-form solution in \cite{umeyama1991least}, and the value of the summation is referred to as ATE of the algorithm's estimate over the dataset. A MATLAB script is also provided in this data suite to perform this task.

\section{Known issues and limitations} \label{sec: issues}
Despite careful design and execution of the data collection experiments, we are aware of several practical issues which pose some challenges and limit the achievable
accuracy. The specifics are given below

\subsection{Camera exposure setting}
Stereo camera exposure setting is a challenging topic for any global-shutter stereo camera. Since the incoming photons to the CMOS sensor can be vastly different at different poses, the obtained image quality can also vary quite dramatically. However, stereo matching needs a set of images with consistent brightness. This dataset uses a set of fixed global shutter exposure settings for various indoor and outdoor locations, but each camera master gain is set to be automatic. Our exposure setting is usually one-third of the default auto-exposure setting for reducing the glare effect. This shorter exposure time improves the image's sharpness and reduces the motion blur at the cost of a slightly darker appearance. The darker appearance can be a challenge for any direct visual SLAM system. We think it is a necessary cost to get a sharp accurate measurement of the environment.

\subsection{Partial loss of information}
In the SBS experiments, the Leica station lost track of the prism for some short periods of time, due to its vantage point being much lower than that in other environments. The evaluation method provided by us already takes this into account in Sec. \ref{sec: resampling}.

Besides, there was significant loss of UWB in some experiments due to low received signal strength (RSS) at certain relative position between the ranging node and the anchor. This in turn is due to the shape of the radiation pattern (see \cite{chen2018impact} for this pattern). This is not the ideal case for a dataset, but it is a real-scenario that the researcher has to address when developing new algorithms.

\section{Conclusion and future works} \label{sec: conclusion}
In this paper we present the datasets collected from a sensor suite typical of autonomous driving car, but equipped on a drone. The datasets are aimed at boosting the investigations into autonomous navigation of UAVs using the up-to-date technologies, which can facilitate progress in many important industries.

Despite our best efforts, there remain some issues in the development of the datasets. However, we do plan to keep our work continuously updated, and these issues will be addressed in the future. For example, the loss of UWB can be remedied by updating the UWB hardware, the tracking of ground truth can be improved with a better vantage point. \highlight{Moreover, temporal alignment can be further improved by batch optimization methods}. The camera hardware can also be made up-to-date with other more popular RGBD-type sensors. Finally, more datasets will be added to the suite in the future.

\section*{Dataset access}
The dataset can be accessed at our web page \url{https://ntu-aris.github.io/ntu_viral_dataset/}
. Other details on calibration, evaluation and updates are also presented at this site to facilitate the use of our dataset on individual's research.

\section*{Funding acknowledgement}
This work is supported in part by the Wallenberg AI, Autonomous Systems and Software Program (WASP), funded by the Knut and Alice Wallenberg Foundation, under the Grant Call 10013 - Wallenberg-NTU Presidential Postdoctoral Fellowship 2020.

\balance
\bibliographystyle{SageH}

\end{document}